\documentclass{article}

\usepackage{arxiv}

\usepackage{graphicx}
\usepackage[list=true, font=small, labelfont=bf, 
labelformat=brace, position=top]{subcaption}
\usepackage{dblfloatfix}

\usepackage{tabularx}
\usepackage{makecell}
\usepackage{multirow}

\usepackage{hyperref}       
\usepackage{url}            
\usepackage{float}
\usepackage{enumitem}
\usepackage{footmisc}

\usepackage{algorithm} 
\usepackage[noend]{algpseudocode} 
\usepackage{amsmath,amsfonts}
\usepackage{bm} 

\setlist[description]{leftmargin=\parindent,}

\DeclareMathOperator{\D}{D}

\DeclareMathOperator{\DPQ}{DPQ}

\newcommand{\overbar}[1]{\mkern 1.5mu\overline{\mkern-1.5mu#1\mkern-1.5mu}\mkern 1.5mu}

\begin{document}
\title{Creating Sorted Grid Layouts with Gradient-based Optimization}

\author{ 
    \href{https://orcid.org/0000-0001-6309-572X}{\includegraphics[scale=0.06]{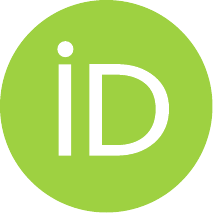}\hspace{1mm}Kai Uwe Barthel} \\
	Visual Computing Group\\
	HTW Berlin\\
	Berlin, Germany \\
	\texttt{barthel@htw-berlin.de} \\
	\And
    \href{https://orcid.org/0000-0001-6309-572X}{\includegraphics[scale=0.06]{orcid.pdf}\hspace{1mm}Florian Tim Barthel} \\
	Fraunhofer HHI\\
	Berlin, Germany \\
        HU Berlin\\
	Berlin, Germany \\
	\texttt{florian.tim.barthel@hhi.fraunhofer.de} \\
	\And
    \href{https://orcid.org/0000-0001-8378-4805}{\includegraphics[scale=0.06]{orcid.pdf}\hspace{1mm}Peter Eisert} \\
	Fraunhofer HHI\\
	Berlin, Germany \\
        HU Berlin\\
	Berlin, Germany \\
	\texttt{peter.eisert@hhi.fraunhofer.de} \\
	\And
    \href{https://orcid.org/0000-0002-3957-4672}{\includegraphics[scale=0.06]{orcid.pdf}\hspace{1mm}Nico Hezel} \\
	Visual Computing Group\\
	HTW Berlin\\
	Berlin, Germany \\
	\texttt{hezel@htw-berlin.de} \\
	\And
    \href{https://orcid.org/0000-0003-3548-0537}{\includegraphics[scale=0.06]{orcid.pdf}\hspace{1mm}Konstantin Schall} \\
	Visual Computing Group\\
	HTW Berlin\\
	Berlin, Germany \\
	\texttt{konstantin.schall@htw-berlin.de} 
}

\maketitle              
\begin{abstract}
Visually sorted grid layouts provide an efficient method for organizing high-dimensional vectors in two-dimensional space by aligning spatial proximity with similarity relationships. This approach facilitates the effective sorting of diverse elements ranging from data points to images, and enables the simultaneous visualization of a significant number of elements. 
However, sorting data on two-dimensional grids is a challenge due to its high complexity. Even for a small 8-by-8 grid with 64 elements, the number of possible arrangements exceeds $1.3 \cdot 10^{89}$ - more than the number of atoms in the universe - making brute-force solutions impractical.

Although various methods have been proposed to address the challenge of determining sorted grid layouts, none have investigated the potential of gradient-based optimization. In this paper, we present a novel method for grid-based sorting that exploits gradient optimization for the first time. 
We introduce a novel loss function that balances two opposing goals: ensuring the generation of a "valid" permutation matrix, and optimizing the arrangement on the grid to reflect the similarity between vectors, inspired by metrics that assess the quality of sorted grids. 
While learning-based approaches are inherently computationally complex, our method shows promising results in generating sorted grid layouts with superior sorting quality compared to existing techniques.
\end{abstract}

\section{Introduction}

Viewing large volumes of images is a cognitive challenge for humans, as the perception of certain content decreases as the volume increases. 
Typically, applications and websites display only a small fraction of the available images, about 20 at a time. 
However, sorting images by the similarity of their visual feature vectors allows hundreds of images to be viewed simultaneously, helping users identify regions of interest more efficiently \cite{Schoeffmann2011, Quadrianto2010, journals/tog/ReinertRS13, journals/corr/HanZLXSM15}. 
Visually sorted grid layouts offer advantages, especially for stock agencies and e-commerce platforms, where navigation through the collections is crucial.

The visual feature vectors can be generated by image analysis methods or deep neural networks \cite{Babenko2014, DeepFeatures, Radenovic2019, Cao2020}. 
These vectors have tens of dimensions for low-level features, while deep learning vectors can have thousands of dimensions.

There are many techniques for visualizing high-dimensional data relationships in two dimensions, such as Principal Component Analysis (PCA), Multidimensional Scaling (MDS), t-Distributed Stochastic Neighborhood Embedding (t-SNE) and UMAP \cite{Sarveniazi2014, Maaten2008, McInnes2018}. 
However, these methods all have their limitations when images are to be displayed, as they overlap and do not use the display area efficiently.

In order to arrange the images according to their similarity and at the same time optimize the used display area, the images have to be arranged in a rectangular grid, where the spatial distances between the images should be as close as possible to the distances of the high-dimensional feature vectors.

While the Traveling Salesman Problem (TSP) focuses on finding the most efficient path between vectors corresponding to their arrangement on a one-dimensional grid, the challenge significantly increases when considering 2D grid layouts. In this scenario, there are still $n!$ ways to arrange the vectors. However, unlike the TSP, where only the similarity to the previous and next neighbor vector matters, the 2D grid layout introduces a more complicated constellation. Due to the two-dimensional grid structure, each vector has four neighbors that are required to be as similar as possible.

Various techniques for generating grid-based layouts are discussed further in Section \ref{sec:related_work}.

While several metrics exist for evaluating sorted arrangements, research on the correlation between human-perceived quality and metric values is limited. 
To address this gap, we introduced Distance Preservation Quality (DPQ) as a more appropriate metric for evaluating arrangement quality, which shows a stronger correlation with user-perceived sorting quality of grid layouts than other metrics \cite{Improved_Wiley}.
\\
\\
In machine learning, many problems involve learning the optimal permutation of a set of objects. This includes tasks such as optimal assignment problems, where objects must be matched based on certain criteria; ranking lists, which involve ordering objects according to preference or importance; and sorting numbers, where the goal is to arrange data in a specific sequence
\cite{sinkhorn1964relationship, santa2018visual, petersen2022monotonic}. 

The sorted arrangement of vectors in a 2D grid can also be described by a permutation matrix, which is a square binary matrix that has exactly one entry of 1 and all other entries of 0 in each row and each column.
However, learning in these models is difficult because exact calculation over these combinations is impractical. A common approach is the use of a convex relaxation of the permutation matrix into a doubly stochastic matrix.

The key contributions of our work are: 
\begin{enumerate}
    \item We introduce a differentiable grid-based sorting scheme and define a suiting loss function to ensure convergence to a permutation matrix while preserving spatial similarity.
    \item We evaluate different techniques to create differentiable permutation matrices.
    \item We achieve a sorting quality that exceeds the state-of-the-art.
\end{enumerate}

\section{Related Work} \label{sec:related_work}

Grid-based sorting of vectors can be approached as a Quadratic Assignment Problem (QAP) \cite{Koopmans}, where the objective is to optimally place $n$ facilities into $n$ distinct positions such that the total transportation cost is minimized. 
This cost is determined by the flow between facilities and the distance between their assigned positions. Specifically, there is a transportation demand (or flow) of $f_{ij} \geq 0$ between facilities $i$ and $j$, while the distance between their assigned positions, as determined by a permutation $\phi$, is given by $d_{\phi(i)\phi(j)} > 0$.

To apply the Quadratic Assignment Problem to the grid-based sorting problem, a mapping between vector distances, spatial grid distances, and the flows and distances described in the QAP is needed. In grid-based sorting, vectors mapped close together should be as similar as possible.  This requirement leads to a reciprocal relationship between the flows and grid distances.
This can be achieved by treating vector distances as the flow measure and using a similarity function based on grid position proximity instead of distance.

Despite considerable research, the Quadratic Assignment Problem remains one of the most complex optimization problems, particularly when compared to the simpler Linear Assignment Problem (LAP). No exact algorithm can efficiently solve QAP instances larger than size $n = 20$ within a reasonable computational time, which has led to the use of approximation methods for grid-based sorting. Some of these methods are briefly discussed below.

\subsection{Distance Preserving Grid Layout Algorithms}
Distance Preserving Grid Layouts reorganize the assignment to the grid positions while trying to align spatial proximity with similarity relationships of the vectors. In the following, the most common approaches are briefly described.
\\
\textbf{Grid Arrangements} 
\\
A \textit{Self-Organized Map} (SOM) \cite{Kohonen1982, Kohonen2013} employs unsupervised learning to generate a lower-dimensional, discrete representation of the input space. It consists of a rectangular grid of map vectors, each matching the dimensionality of the input vectors. The input vectors are assigned to the most similar map vectors, which are updated by their neighborhood after assignment. When adapting a SOM for image sorting, each input vector must be assigned to a unique  best-fitting map position. To avoid image overlap the number of map positions must be at least as large as the number of images to be sorted.

A \textit{Self-Sorting Map} (SSM) \cite{Strong2011,Strong2014} initializes cells with the input vectors. It then uses a hierarchical swapping process, comparing the similarity of vectors to the average of their grid neighborhood, considering all 24 possible swaps for sets of four cells.

\label{Linear Assignment Sorting}
Our previously proposed \textit{Linear Assignment Sorting} (LAS) \cite{Improved_Wiley} combines ideas from the SOM (using a continuously filtered map) with the SSM (swapping of cells) and extends this to optimally swapping all vectors simultaneously. \textit{Fast Linear Assignment Sorting} (FLAS) enhances runtime efficiency while achieving only slightly reduced sorting quality compared to LAS.  
\\
\\
\textbf{Graph Matching} 
\\
\textit{Kernelized Sorting} (KS) \cite{Quadrianto2008} and \emph{Convex Kernelized Sorting} \cite{Djuric2012} generate distance-preserving grids and seek a locally optimal solution to a quadratic assignment problem \cite{Beckman1957}. KS computes a matrix of pairwise distances between the high-dimensional (HD) data instances and another matrix between grid positions. It then modifies the latter through a permutation procedure to approximate the former, resulting in a one-to-one mapping between instances and grid cells.

\textit{IsoMatch} also employs an assignment strategy to construct distance-preserving grids \cite{Fried2015}. Initially, it projects data into the 2D plane using the Isomap technique \cite{Tenenbaum2000} and forms a complete bipartite graph between the projection and grid positions. Subsequently, the Hungarian algorithm \cite{Kuhn1955} is utilized to find the optimal assignment for the projected 2D vectors to the grid positions, maximizing overall distance preservation using the normalized energy function.

Any other dimensionality reduction methods (such as t-SNE or UMAP) can be applied to project high-dimensional input vectors onto the 2D plane before rearranging them on the grid. Fast placing approaches are available in \cite{Hilasaca2021}. Linear assignment schemes like the Jonker-Volgenant Algorithm \cite{Jonker1987} can be employed to map projected 2D positions to the best grid positions.


\subsection{Learning Latent Permutations}

Machine learning relies on gradient descent methods, as it typically involves minimizing a differentiable function over parameters. However, in cases where variables are discrete, gradient descent cannot be applied unless these variables are relaxed to continuous ones for which derivatives can be defined. In \cite{46645}, it was shown how permutations can be approximated using the differentiable Sinkhorn operator. Gumbel-Sinkhorn networks were used to solve tasks such as sorting numbers and solving jigsaw puzzles. Other applications are linear and quadratic assignment, and shape matching problems.
The authors define the Sinkhorn operator $S$ on a square matrix $M$ as follows:
\begin{equation}
\begin{array}{l}
    S^0(M) = \exp{(M)} 
\\
S^l(M) = \mathcal{T}_r(\mathcal{T}_c(S^{l-1}(M)))
\\
S(M) = \lim_{l \to \infty}S^l(M)
  \end{array}
\end{equation}

The operations $\mathcal{T}_r$ and $\mathcal{T}_c$ normalize the matrix $M$ by dividing each row by the sum of its elements (for $\mathcal{T}_r$) and each column by the sum of its elements (for $\mathcal{T}_c$). Repeated iterations of this process lead to the generation of a doubly stochastic matrix. The authors introduce the Gumbel-Sinkhorn operator by applying the Sinkhorn operator on an unnormalized assignment probability matrix to which Gumbel noise $\epsilon$ scaled by $\beta$ in the range of 0 to 1 is added. A permutation matrix $P$ can be approximated by the differentiable continuous relaxation
\begin{equation}
    P \approx S(\frac{M+\beta\cdot\epsilon}{\tau}) 
\end{equation}
For small values of the scaling factor $\tau$, the application of the Gumbel-Sinkhorn operator to a matrix M converges to a permutation in the limit.
\\
\\
To generate a permutation, it is essential to define how it can be represented by the weights undergoing learning. Theoretically, $n$ discrete numbers specifying the resulting sequence are sufficient for $n$ elements. The problem is that this approach is not differentiable. 
The description of the full permutation matrix and subsequent learning with a Gumbel-Sinkhorn network requires $n^2$ variables, which leads to significant memory requirements for a large number of elements. To address this problem, Droge et al.~\cite{Droge2023KissingTF} propose to overcome this problem associated with large permutation matrices by approximating them by a low-rank matrix factorization followed by non-linearity.

A reduction to only $n$ weights is theoretically possible using the \textit{SoftSort} method \cite{DBLP:conf/icml/PrilloE20}, which is used for differentiable sorting of variables. However, this approach is not suitable for 2D grid layouts due to the underlying 1D ordering principle.

We evaluate these different approaches in Section \ref{sec:Evaluated_Learnable_Permutations}.
\\
\section{Creating Sorted Grid Layouts with Gradient-based Optimization} 
A significant difference in learning sorted grid layouts lies in the fact that the number of elements to be sorted is much larger compared to other previously proposed permutation learning tasks. While optimal (1D) sorting of numbers cannot be achieved for sets larger than about 100 elements, 2D puzzle tasks become unsolvable even with grid sizes as small as 5x5 elements \cite{46645, petersen2022monotonic}. In grid-based sorting, such as sorting images, one often deals with several hundred elements. Another distinction from number sorting or puzzle tasks is that the optimal solution is unknown for grid-based sorting and therefore cannot be utilized for training.

Grid-based arrangement of the HD vectors $X$ consists of finding a permutation $P: X \to Y$ or $x_i \mapsto y_j$, where $x_i$ is the $i^{\:th}$ high-dimensional vector, whereas $y_j$ is the $j^{\:th}$ position vector on the grid with $n$ elements in $\mathbb{R}^2$. 
The learned permutation should arrange the $X$ vectors in such a way on the 2D grid that their spatial proximity corresponds as closely as possible to their similarity.

\begin{figure}[t]
\centering
\includegraphics[width=0.8\linewidth]{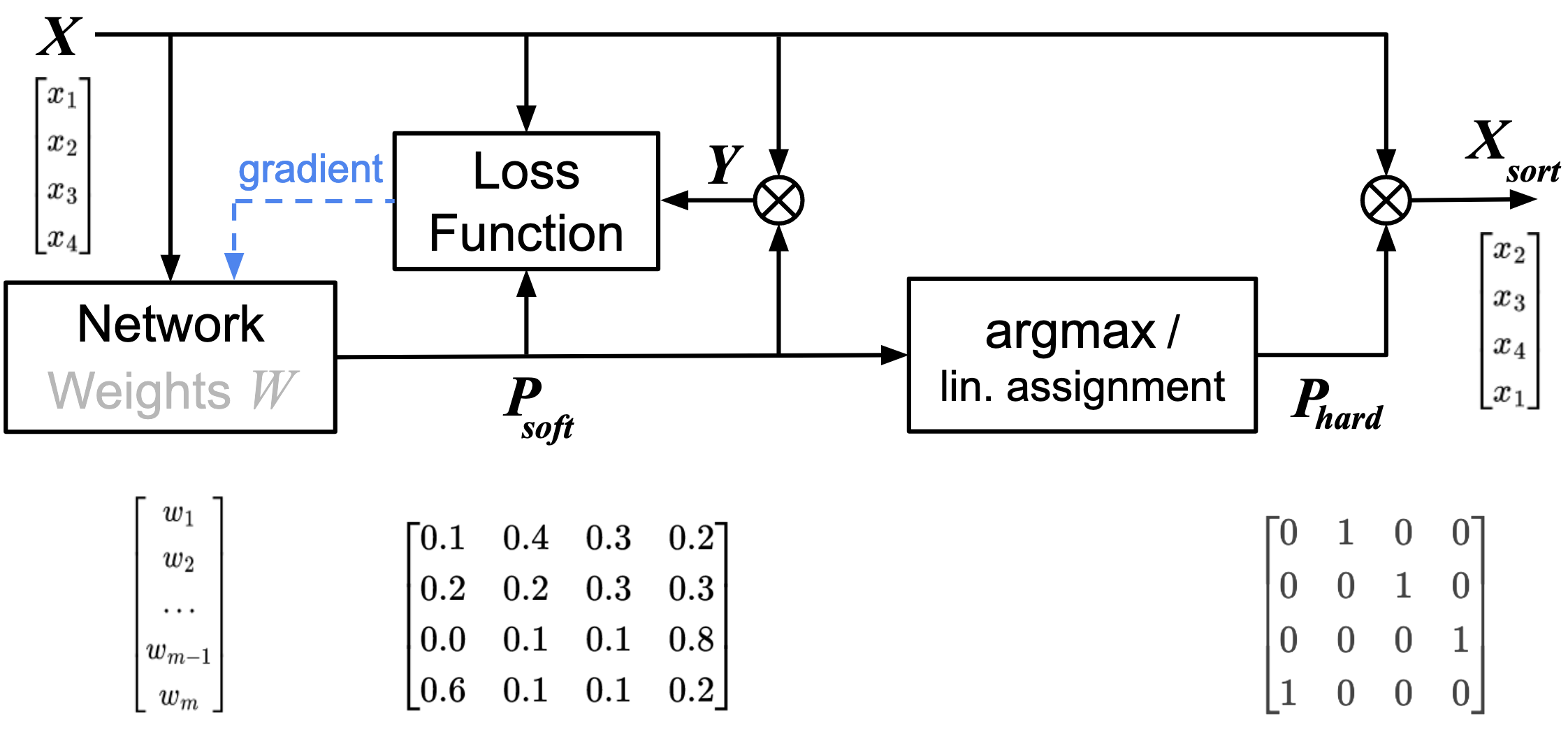}
  \caption{\label{fig:figure1}
The principle of permutation learning involves training a network with weights \textit{W} to learn the differentiable permutation matrix \textit{P\textsubscript{soft}} based on a specified loss function. The input vectors \textit{X} are rearranged into \textit{X\textsubscript{sort}} using the permutation matrix \textit{P\textsubscript{hard}}, which is a binarized version of \textit{P\textsubscript{soft}}.
 } 
\end{figure}

Figure \ref{fig:figure1} illustrates the fundamental process of the permutation learning. A network with weights $W$ generates the matrix $P_{\textit{soft}}$. For the network, different methods for the estimation of permutation matrices can be used as described in Section \ref{sec:Evaluated_Learnable_Permutations}. Initially,  $P_{\textit{soft}}$ is not a permutation matrix because it fails to meet the requirement of having only one '1' per row and column, with all other elements being '0's. The operation $Y=P_{\textit{soft}} \cdot X $ maps the original unsorted vectors $X$ to $Y$. At the beginning of the training process, $Y$ is not yet a reordering of the $X$ vectors but a linear combination of them. 

In order to converge to the permutation matrix, the weights $W$ are learned by optimizing a loss function.
The loss function must satisfy two primary goals: First, it must guarantee that $P_{hard}$, which is a binarized version of $P_{\textit{soft}}$, converges to a "valid" permutation matrix. Second, it should ensure that the final arrangement on the grid is such that it assigns similar $X$ vectors to spatially close positions on the grid. 

The final sorted arrangement is generated using $ X_{\textit{sort}} = P_{\textit{hard}} \cdot X $. 
This operation is valid only if $ P_{\textit{hard}} $ is indeed a permutation matrix. This condition holds if the row-wise $argmax$ of $ P_{\textit{soft}} $ does not contain any duplicates. 
In cases where duplicates are present, $P_{\textit{hard}}$ can still be obtained by establishing a mapping from $P_{\textit{soft}}$ to $P_{\textit{hard}}$ using a linear assignment solver. This solver determines the permutation matrix that minimizes the mean error when mapping $X$ to $Y$.
However, due to the high computational costs of the linear assignment, especially for large $ n $, it is desirable to choose an approach that does not create duplicates and thus avoids the linear assignment.
  
\subsection{Learning Objectives} 
As mentioned before, the inherent non-differentiability of the permutation matrix prohibits its direct learning. Consequently, only $P_{\textit{soft}}$ is learnable. During training, two contradictory goals must be balanced. First, the learning process aims to determine $P_{\textit{soft}}$ in such a way that a valid permutation matrix $P_{\textit{hard}}$ can be generated. Any matrix that contains only one "1" per row and column and zeros otherwise would fulfill this requirement. 
Such a matrix however cannot guarantee the similarity between the grid elements. This similarity requirement would ideally be achieved with identical $Y$ vectors, generated by a constant matrix where each element is $\frac{1}{n}$. 
However, this maximized similarity contradicts the requirements for a valid permutation matrix.
For this reason, the loss function must balance these contradictory requirements. 
\\
\\
In the following, we describe the loss functions for the mapping of the set of vectors $X = \{x_1, x_2, ..., x_n\}$ onto a rectangular grid with $ n = n_x \cdot n_y $ grid positions. 
The squared distances $\delta$ of the vectors $X$ are defined as
\begin{equation}
\delta(x_i, x_j) = \sum_{d}(x_{i,d}-x_{j,d})^2
\end{equation}
where $d$ is the index of the dimension of the vectors. The squared distances $\delta'$ of the sorted vectors $Y$ are defined as
\begin{equation}
\delta'(y_i, y_j) = \sum_{d}(y_{i,d}-y_{j,d})^2
\end{equation}
\noindent The global average squared distance $\overbar{\D}$ of all vectors $x_i$ is

\begin{equation}
\overbar{\D} = \frac{1}{n(n-1)} \sum_{i=1}^{n} \sum_{j=1}^{n} \delta(x_i, x_j)
\end{equation}
\\
To assess the desired smoothness of the mapping, a suitable quality metric must be used. In \cite{Improved_Wiley}, it was shown that the \textit{Distance Preservation Quality} (DPQ) is suitable to evaluate the sorting quality of a 2D grid layout. The problem, however, is that the DPQ calculation is computationally intensive and non-differentiable, which prohibits its use in training.
However, the DPQ metric uses the parameter $ p $ for which optimal values are about 16. For larger $ p $ values, when only the similarity of the closest neighbors is taken into account, the agreement with user ratings remains relatively consistent. For large $p$ values, the behavior of the DPQ metric can be approximated by minimizing the mean square distance of the vectors to their four nearest neighbors on the grid. Therefore, we introduce the \textit{neighborhood loss} $L_{nbr}$ of a matrix $P$ as follows:
\begin{equation}
\D_{hor}(P)=\frac{1}{(n_x-1)n_y}\sum_{\substack{i=1 \\ i \% n_x \neq 0} }^{n}  \delta'(y_{i},y_{i+1}) 
\end{equation}

\begin{equation}
\D_{ver}(P)=\frac{1}{n_x(n_y-1)}\sum_{i=1}^{n-n_x}  \delta'(y_{i},y_{i+n_x}) 
\end{equation}

\begin{equation} \label{eq:8}
L_{nbr}(P)=\frac{\D_{hor} + \D_{ver}}{2 \cdot \overbar{\D}},
\end{equation} 
with $\D_{hor}$ and $\D_{ver}$ being the average distance of neighboring grid vectors in horizontal and vertical direction, respectively. The normalization with the global average squared distance $\overbar{\D}$ results in $L_{nbr}(P)$ being in the range from 0 to 1.

\noindent The neighborhood loss can also be expressed as quadratic assignment problem: 

\begin{equation}  \label{eq:9}
L^*_{nbr}(P) = \frac{1}{ \overbar{\D} \cdot \sum_{i} \sum_{j} sim_{ij}} \sum_{i} \sum_{j} \delta'_{ij} \cdot sim_{ij}
\end{equation}
where $sim_{ij}$ represents the spatial similarity or proximity based on the 2D distance of the grid positions $d^{2D}_{ij}$, ranging from 1 for the closest neighbors to 0 for the most distant neighbors on the grid: 
\begin{equation}
sim_{ij} = \Biggl(\frac{d^{2D}_{max} - d^{2D}_{ij} }{d^{2D}_{max} -1} \Biggr)^p, \quad sim_{ii} = 0
\end{equation}
The parameter $p$ determines the degree to which the similarity values decrease with increasing spatial distance. 
This allows to control the extent to which more distant neighbors are to be included in the loss term.
For higher $p$ values, the loss from Equation (\ref{eq:9}) approaches that of Equation (\ref{eq:8}), since all $sim$ values except those of the nearest direct neighbors will be 0.
\\
\\
Several methods can be employed to guide the learned matrix toward converging to a permutation matrix. 
In \cite{10.1145/3394486.3403103}, the $L1\_2$ \textit{penalty} is introduced, which contrasts the row-wise $L1$ and $L2$ norms of a stochastic matrix. The discrepancy is minimal, approaching 0, when the matrix is a permutation matrix. We have found that the $L1\_2$ penalty - depending on its weighting - either does not lead to permutation matrices or leads to bad arrangements.

In our investigations, we obtained the best results by combining two loss components to generate valid permutation matrices. 
The first regularization term, the \textit{stochastic constraint loss} $L_s(P)$, ensures that $P_{\textit{soft}}$ converges to a doubly stochastic matrix by penalizing deviations from 1 in the row or column sums of the matrix:
\begin{equation}
L_s(P)=\frac{1}{n}\sum_i \Bigl( \sum_j |P_{ij}|-1  \Bigr)^2 + \frac{1}{n}\sum_j \Bigl( \sum_i |P_{ij}|-1  \Bigr)^2 
\end{equation}

The next step is not to check directly whether the matrix contains only one "1" per row and column and zeros elsewhere, but rather to compare the distance matrices $D_X$ and $D_Y$ of the vectors $X$ and $Y$. 
\begin{equation}
D_X = (\delta(x_i, x_j)) \;\;\;\;\;\;  D_Y = (\delta'(y_i, y_j)) 
\end{equation}

Since the vectors $X$ and $Y$ have different orders of their elements, they can only be compared if both distance matrices are sorted row-wise and column-wise. This leads to the second regularization term the \textit{distance matrix loss} $L_p(P)$ to enforce $P_{\textit{soft}}$ to converge to a permutation matrix.
\begin{equation}
L_p(P) = \frac{ \sum_{ij}|sort_r(sort_c(D_X)) - sort_r(sort_c(D_Y))|}{\sum_{ij}D_X}
\end{equation}
$sort_r$ and $sort_c$ perform row- and column-wise sorting of the elements of both distance matrices $D_X$ and $D_Y$.
\\
\\
The final loss function $L(P)$ used in our neural grid-based sorting scheme is:
\begin{equation} \label{eq:loss}
L(P) = \underbrace{L_{nbr}(P)}_{\text{smoothness term}} + \;\;\; \underbrace{\lambda_s L_s(P) + \alpha(t)\cdot \lambda_p L_p(P)}_{\text{regularization terms}} 
\end{equation}
$\alpha(t)$ represents a weighting factor that ranges from 0 at the beginning of training to 1 at the end. 
We have found that it is advantageous to gradually include the distance matrix loss $L_p(P)$ to prevent a too fast convergence to a non-smooth mapping / permutation. The parameters $\lambda_s$ and $\lambda_p$ control the strength of the regularization terms. 
\\
\subsection{Evaluated Learnable Permutations} \label{sec:Evaluated_Learnable_Permutations}

\begin{figure}[t]
\centering
\includegraphics[width=1\linewidth]{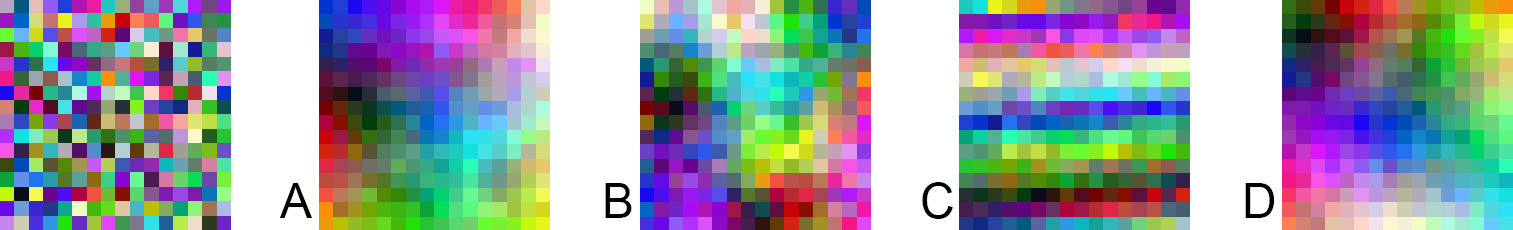}
  \caption{\label{fig:color_example}
An example with 256 RGB colors. On the left, the colors are randomly arranged. Sorted examples using different methods to generate the permutation matrix: A) Full rank + Gumbel-Sinkhorn, B) Low-rank + Softmax, C) SoftSort  with \textit{n} weights (achieving only 1D sorting), D) Full rank, Transformer + Gumbel-Sinkhorn. 
 } 
\end{figure}

Several methods can be used to generate the differentiable permutation matrix $P_{\textit{soft}}$ within the network (as depicted in Figure \ref{fig:figure1} on the left). Below, we outline four methods. Figure \ref{fig:color_example} shows examples of sorting results for a set of 256 random colors, demonstrating the outcomes of these approaches.
\\
\\
\noindent \textbf{A) Gumbel-Sinkhorn}
\\
 The Gumbel-Sinkhorn method \cite{46645} allows to  approximate  
 the permutation matrix from a full rank matrix with $n^2$ parameters. There is no neural network involved, the weights are directly fed to the $n \times n$ matrix to which the Gumbel-Sinkhorn operator is applied iteratively.  
\\
\\
\noindent\textbf{B) Low-Rank Permutation Representation}

\noindent Using matrix factorization \cite{Droge2023KissingTF} reduces the number of parameters from $n^2$ to $2nm$ with  $m \ll n$. Given two row-normalized matrices $V$ and $W$ in $\mathbb{R}^{n \times m}$, the elements of $VW^T$ represent the cosines of the angles between the rows of the matrices. Applying the row-wise Softmax to the product of the matrices enables the representation of any permutation. The kissing number theory is used to estimate the minimum realizable value of $m$. 
In our investigations, we found that matrix factorization usually yielded poorer permutations. 
Furthermore, the advantage of parameter reduction is negligible for these small quantities of vectors.
\\
\\
\noindent \textbf{C) SoftSort}
\\
A theoretical approach to modeling the permutation matrix with only $n$ parameters can be explored with \textit{SoftSort} \cite{DBLP:conf/icml/PrilloE20}, a relaxation of the argsort operator that returns the sorting permutation of an input vector. Using the described loss function, one can try to achieve grid-based 2D sorting with SoftSort. However, the inherent ordering mechanism of SoftSort inevitably leads to 1D sorting in practice.
\\
\\
\noindent \textbf{D) Transformer Network \& Gumbel-Sinkhorn}

\noindent The previously described methods create permutation matrices from learnable weights without any information of the input vectors $X$. Since the loss function computes gradients w.r.t.~$X$, those weights will be optimized to yield permutations with a high sorting quality. We explored the idea of integrating $X$ to the permutation generation process with the help of a Transformer architecture \cite{Attention} and follow the implementation as used for vision transformers (ViT) \cite{VIT}. 
This addition could be applied to all of the previously described methods. First, each element $x_i$ is projected to a desired embedding dimensionality $d$ with a linear layer. Each projected $x_i$ now represents a single \textit{token} $t_i$. Next, the sequence of tokens is prepended with a learnable vector in $\mathbb{R}^{d}$. In vision transformers this vector is often used as a \textit{class token}, however, in our case this token will not be used for classification but rather to produce the desired weight matrix, which will then be transformed into a permutation matrix. The entire sequence of tokens is now put through a number of \textit{Multi-Head Self-Attention} layers \cite{Attention}, which will output a transformed sequence of tokens with length $n+1$. The transformed \textit{class token} is projected to a vector with the desired dimensionality in a last step, which is $n^2$ if the Gumbel-Sinkhorn method is to be used subsequently. This approach introduces high computational costs. Since the Multi-Head Attention layers have quadratic complexity, this method is not suitable for sorting grids with a very large number of elements.

\section{Experiments}
In our experiments, we evaluate the performance of different state-of-the-art methods for grid-based sorting. We compare these methods with two implementations of our novel gradient-based sorting method, namely \textit{GradSort} and \textit{GradSort Transformer}, which are based on methods A and D presented in the last section.

The data points to be sorted, whether images or colors, are represented by their respective feature vectors.
Based on the test protocol described in \cite{Improved_Wiley}, the following datasets\footnote{https://github.com/Visual-Computing/LAS\_FLAS/blob/main/data.zip} are used in this section: 
Traffic Signs (169 images), Kitchenware (256 images), Web Images (400 images), and a set consisting of 1024 random RGB colors. 
These datasets were chosen for their varying levels of complexity and relevance to real-world sorting tasks. 
The image datasets were processed using low-level image analysis techniques to extract 50-dimensional feature vectors representing the most important visual properties for effective sorting \cite{Improved_Wiley}. Such low-level feature vectors are better suited for the visual sorting of larger image sets, as they allow humans to visually group similar images more easily. These vectors are discriminative enough to ensure that images within the same category have similar features, preventing suboptimal sortings due to inadequate feature vectors. The RGB colors are represented by their red, green, and blue values.

To determine the sorting quality, the \textit{Distance Preservation Quality} metric $\DPQ_{16}$ was used, which emphasizes close spatial neighborhood of similar vectors / images. This metric was chosen due to its proven correlation with the perceived quality of arrangements by humans \cite{Improved_Wiley}. The time required for sorting was recorded, but only serves as an approximate indicator. 

For our newly proposed method, all networks were trained with identical learning parameters for all datasets to ensure a fair comparison. These parameters were selected following initial experiments. 
The training was performed using the Adam optimizer with a learning rate of 0.03. For the Gumbel-Sinkhorn operator 10 iterations were executed using a $\tau$ value of 1. The factor $\beta$ for adding Gumbel noise was set to 0.1. In order to generate sortings of different quality, the training was carried out with different maximum training steps, which leads to smaller $\alpha$ values for increasing numbers of training steps (Equation (\ref{eq:loss})). The training continued until there were no more duplicates when the permutation matrix was created with the argmax operator. 
Typically, about half of the maximum training steps were required until convergence, while the number was further reduced when the transformer was used.
The regularization parameters $\lambda_p$ and $\lambda_s$ of the loss function were set to 5 and 100 respectively. The utilized Transformer network is rather small and uses only three Multi-Head Self-Attention layers with a hidden dimensionality of 16 and 4 heads.
For the other compared methods, a search was conducted for optimal hyperparameters to find the settings that maximize the $\DPQ_{16}$ value.

The testing environment consisted of a Ryzen 2700x CPU with a fixed core clock of 4.0 GHz and 64GB of DDR4 RAM running at 2133 MHz. All experiments were executed on a single CPU thread. 
The new \textit{GradSort} algorithm was implemented in Python, while the other algorithms in either Java or Octave. Further information about the setup and testing methods can be found at \cite{Improved_Wiley}. 
At startup, all data was loaded into memory. Then the averaged runtime and $\DPQ_{16}$ value of 20 runs were recorded. 
Different initial data orders were used for the individual runs. However, we ensured that all algorithms received the same corresponding orders.

\bgroup
\def\arraystretch{1.2}
\begin{table}[h]
    \centering
    \begin{tabularx}{\columnwidth} 
    { |
   >{\centering\arraybackslash}l  |
   >{\centering\arraybackslash}X  |
   >{\centering\arraybackslash}X  |
   >{\centering\arraybackslash}X  |
   >{\centering\arraybackslash}X  |}
        \hline
        \multirow{2}{*}{
        \parbox{2.1cm}
        {  \centering
        \ \\  [-0.9ex]
        \hspace{2.1em} \textbf{Dataset}
        \  \\ [0.8ex]
        \hspace{-2.6em} \textbf{Algorithm}
        }
        } &
        \multicolumn{1}{c|}{\textbf{Color}} &
        \multicolumn{1}{c|}{\textbf{Traffic}} &
        \multicolumn{1}{c|}{\textbf{Kitchen-}} &
        \multicolumn{1}{c|}{\textbf{Web}} \\ 
        \cline{1-1}
        & \textbf{1024}
        & \textbf{Signs} 
        & \textbf{ware} 
        & \textbf{Images} \\
        \hline
        IsoMatch    & 0.783 & 0.810 & 0.785 & 0.741 \\
        SOM         & 0.918 & 0.858 & 0.853 & 0.856 \\
        SSM         & 0.924 & 0.880 & 0.873 & 0.858 \\
        t-SNEtoGrid & 0.917 & 0.905 & 0.904 & 0.905 \\
        FLAS        & 0.945 & 0.919 & 0.888 & 0.892 \\
        LAS & \textbf{0.954} & 0.919 & 0.896 & 0.896 \\
        \hline       
        GradSort (\textbf{ours}) & \textbf{0.954} & \textbf{0.936} & \textbf{0.931} & \textbf{0.928} \\
        \hline
    \end{tabularx}
    \vspace{0.2cm}
    \caption{Comparison of grid-based sorting methods for different test sets based on maximum achievable average DPQ\textsubscript{16} value (higher is better). Our new GradSort scheme was trained for up to 100,000 steps.
    }
    \label{tab:results}
\end{table}
\egroup



\begin{figure}[t]
\centering
\includegraphics[width=0.7\linewidth]{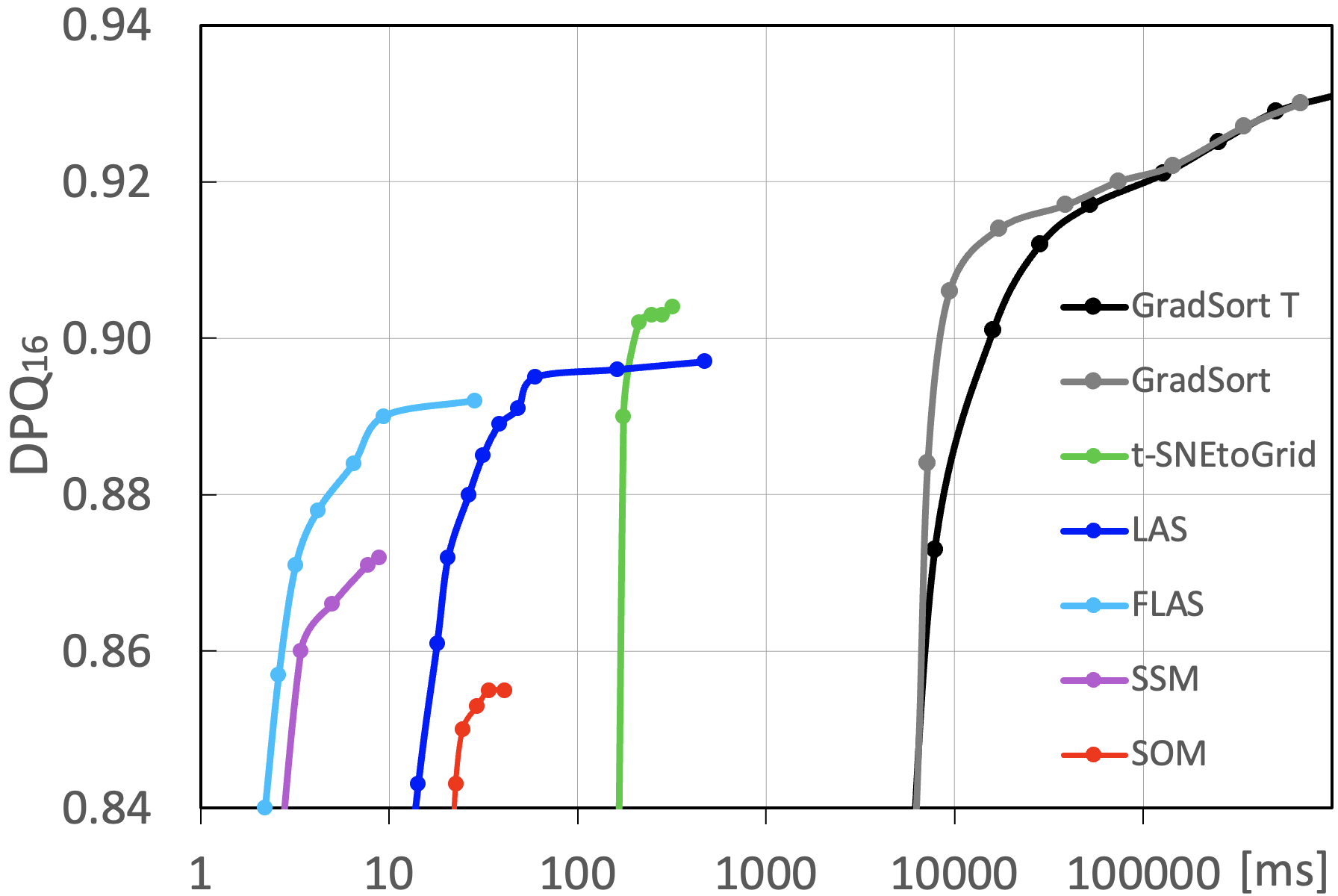}
  \caption{\label{fig:kitchen_runtime}
Comparison of the runtime and the achieved sorting quality of different sorting algorithms with the kitchenware image set. \textit{GradSort T} uses the additional transformer. 
 } 
\end{figure}

\noindent Table \ref{tab:results} presents the maximum average $\DPQ_{16}$ values achieved by different algorithms across various datasets. Figure \ref{fig:kitchen_runtime} illustrates the relationship between runtime and sorting quality for the kitchenware dataset, with similar patterns observed across other datasets.
\textit{GradSort} outperforms state-of-the-art methods in sorting quality, though it requires more time to run compared to other methods. 

While methods like SSM and FLAS offer fast sorting speeds, they compromise on quality. Conversely, LAS balances relatively high $\DPQ_{16}$ values with moderate runtimes.
GradSort significantly improves sorting quality across three image datasets — traffic signs, kitchenware, and web images — while achieving state-of-the-art results with the Colors 1024 dataset.

These results suggest that traditional sorting methods are effective for low-dimensional data, such as three-dimensional RGB values, but struggle with high-dimensional data like feature vectors. However, our proposed method is more robust to data dimensionality, consistently producing high DPQ values across all datasets.

\begin{figure*} 
\centering
\includegraphics[width=1\textwidth]{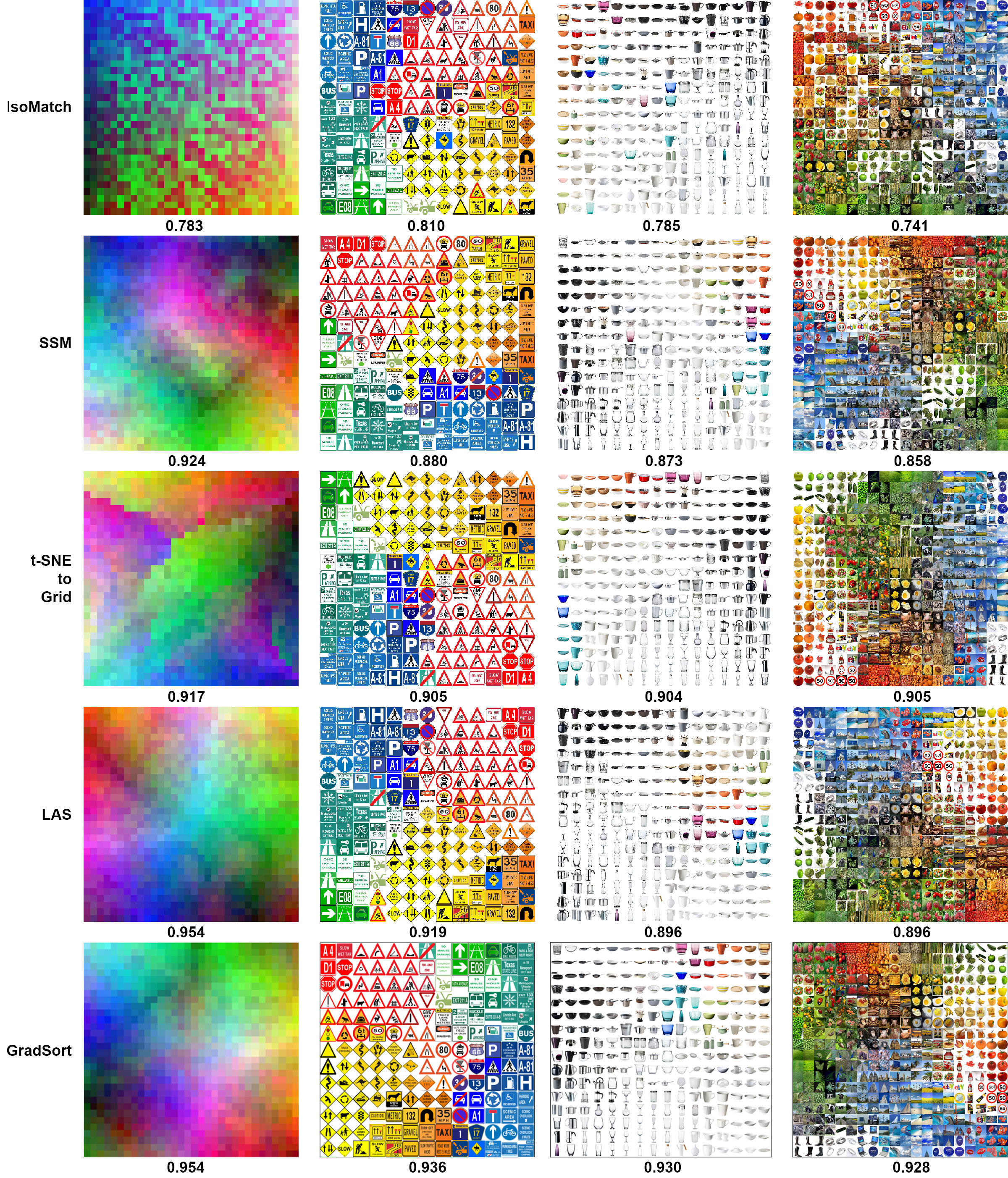}
\caption{A comparison of different sorting methods for the color dataset and three image sets. The top row shows the IsoMatch sorting, followed by the Self Sorting Map (SSM). The third row shows the result of a t-SNE projection, mapped onto the grid positions using a linear assignment solver.  The fourth row displays the results of Linear Assignment Sorting (LAS). The final row exhibits sortings from our new approach,  \textit{GradSort}. Achieved DPQ\textsubscript{16} values are indicated below the respective arrangements. 
}
\label{fig:comparison}
\end{figure*}

The visual assessment, as depicted in Figure \ref{fig:comparison}, highlights a distinct contrast between projection methods such as IsoMatch and t-SNE, and grid-based sorting techniques like SSM, LAS, and \textit{GradSort}. Projection methods begin by mapping data points to a 2D space and then use solvers to arrange these potentially overlapping points into a grid layout. Since the number of grid cells corresponds to the number of data points, solvers often have difficulties in creating optimal mappings, which leads to pronounced discontinuities or dithering effects. When comparing the sorted arrangements of the image datasets across the different sorting methods, examining images that represent the same visual concept can be helpful. The Web Images dataset in the last column contains 70 concepts, each with three to six images. It can be seen that \textit{GradSort} places images with the same concept much closer to each other than LAS.

We anticipated that GradSort Transformer would outperform the simpler GradSort approach which relies solely on weight parameters transformed using the Gumbel-Sinkhorn operator. While the transformer version did converge more quickly, its significantly higher computational complexity negated any performance advantage. This is illustrated in Figure \ref{fig:kitchen_runtime}.

We also explored the impact of including distant neighbors in the neighbor loss function according to Equation (\ref{eq:9}) to improve the sorting quality. However, this approach proved beneficial only for lower quality sortings. For the highest quality sorting, using Equation (\ref{eq:8}) consistently produced better results.

\section{Conclusion} 

The contributions of this paper can be summarized as follows: First, we explore the novel concept of using gradient-based learning for differentiable permutation matrices to create 2D image arrangements. Second, we propose a combination of loss objectives to facilitate gradient descent. Third, we conduct experiments with various methods for generating permutation matrices.
The combination of the Gumbel-Sinkhorn approach and the proposed learning algorithm achieves very high sorting quality, surpassing traditional methods such as SOM, SSM, FLAS, LAS and projections schemes like IsoMatch or a t-SNE projection mapped to a grid.

However, the current approach is significantly slower compared to previous methods due to the iterative nature of the learning procedure and requires GPU-based computations, especially if the number of images is very high or if a large Transformer network is added. 
Future research should focus on optimizing the learning algorithm, such as reducing the number of iterations needed. Although the network typically converges quickly to a high-quality arrangement, most of the remaining time is spent minimizing duplicate assignments. Refining the linear curve for the alpha value controlling regularization could potentially enhance efficiency.

A further improvement when using very high dimensional vectors could be to no longer work directly with the vectors but to use only their distance matrices.

Moreover, the test sets evaluated in our study are relatively small in size. Consequently, methods that reduce the number of parameters, such as matrix factorization, offer limited benefits in these scenarios but could play a more significant role with larger image datasets. We have succeeded in generating a matrix factorization based on the twofold application of SoftSort with only $2n$ parameters, but the sorting results are not yet as good as those presented here.

Research would benefit from standardized large image collections with feature vectors for more comparable and comprehensive experiments.

%
%
%
\bibliographystyle{unsrt}
\bibliography{references}

\begin{thebibliography}{10}

\bibitem{Schoeffmann2011}
Klaus Schoeffmann and David Ahlstrom.
\newblock Similarity-based visualization for image browsing revisited.
\newblock In {\em 2011 IEEE International Symposium on Multimedia}, pages 422--427, 2011.

\bibitem{Quadrianto2010}
Novi Quadrianto, Kristian Kersting, Tinne Tuytelaars, and Wray~L. Buntine.
\newblock Beyond 2d-grids: a dependence maximization view on image browsing.
\newblock In James~Ze Wang, Nozha Boujemaa, Nuria~Oliver Ramirez, and Apostol Natsev, editors, {\em Multimedia Information Retrieval}, pages 339--348. ACM, 2010.

\bibitem{journals/tog/ReinertRS13}
Bernhard Reinert, Tobias Ritschel, and Hans-Peter Seidel.
\newblock Interactive by-example design of artistic packing layouts.
\newblock {\em ACM Trans. Graph.}, 32(6):218:1--218:7, 2013.

\bibitem{journals/corr/HanZLXSM15}
Xintong Han, Chongyang Zhang, Weiyao Lin, Mingliang Xu, Bin Sheng, and Tao Mei.
\newblock Tree-based visualization and optimization for image collection.
\newblock {\em CoRR}, abs/1507.04913, 2015.

\bibitem{Babenko2014}
Artem Babenko, Anton Slesarev, Alexander Chigorin, and Victor~S. Lempitsky.
\newblock Neural codes for image retrieval.
\newblock In David~J. Fleet, Tomás Pajdla, Bernt Schiele, and Tinne Tuytelaars, editors, {\em Proc. ECCV}, volume 8689 of {\em Lecture Notes in Computer Science}, pages 584--599, Zurich, Switzerland, 2014. Springer.

\bibitem{DeepFeatures}
{Richard Yi} Zhang, Phillip Isola, {Alexei A.} Efros, Eli Shechtman, and Oliver Wang.
\newblock The unreasonable effectiveness of deep features as a perceptual metric.
\newblock In {\em Proceedings - 2018 IEEE/CVF Conference on Computer Vision and Pattern Recognition, CVPR 2018}, Proceedings of the IEEE Computer Society Conference on Computer Vision and Pattern Recognition, pages 586--595. IEEE Computer Society, December 2018.

\bibitem{Radenovic2019}
Filip Radenovic, Giorgos Tolias, and Ondrej Chum.
\newblock Fine-tuning cnn image retrieval with no human annotation.
\newblock {\em IEEE Trans. Pattern Anal. Mach. Intell.}, 41(7):1655--1668, 2019.

\bibitem{Cao2020}
Bingyi Cao, André Araujo, and Jack Sim.
\newblock Unifying deep local and global features for image search.
\newblock In Andrea Vedaldi, Horst Bischof, Thomas Brox, and Jan-Michael Frahm, editors, {\em ECCV (20)}, volume 12365 of {\em Lecture Notes in Computer Science}, pages 726--743. Springer, 2020.

\bibitem{Sarveniazi2014}
Alireza Sarveniazi.
\newblock An actual survey of dimensionality reduction.
\newblock {\em American Journal of Computational Mathematics}, 04:55--72, 01 2014.

\bibitem{Maaten2008}
Laurens van~der Maaten and Geoffrey Hinton.
\newblock Visualizing data using {t-SNE}.
\newblock {\em Journal of Machine Learning Research}, 9:2579--2605, 2008.

\bibitem{McInnes2018}
Leland McInnes and John Healy.
\newblock Umap: Uniform manifold approximation and projection for dimension reduction.
\newblock {\em CoRR}, abs/1802.03426, 2018.

\bibitem{Improved_Wiley}
Kai~Uwe Barthel, Nico Hezel, Klaus Jung, and Konstantin Schall.
\newblock Improved evaluation and generation of grid layouts using distance preservation quality and linear assignment sorting.
\newblock {\em Computer Graphics Forum}, 42(1):261--276, 2023.

\bibitem{sinkhorn1964relationship}
Richard Sinkhorn.
\newblock A relationship between arbitrary positive matrices and doubly stochastic matrices.
\newblock {\em The annals of mathematical statistics}, 35(2):876--879, 1964.

\bibitem{santa2018visual}
Rodrigo Santa~Cruz, Basura Fernando, Anoop Cherian, and Stephen Gould.
\newblock Visual permutation learning.
\newblock {\em IEEE transactions on pattern analysis and machine intelligence}, 41(12):3100--3114, 2018.

\bibitem{petersen2022monotonic}
Felix Petersen, Christian Borgelt, Hilde Kuehne, and Oliver Deussen.
\newblock Monotonic differentiable sorting networks, 2022.

\bibitem{Koopmans}
Tjalling Koopmans and Martin~J. Beckmann.
\newblock Assignment problems and the location of economic activities.
\newblock Cowles Foundation Discussion Papers~4, Cowles Foundation for Research in Economics, Yale University, 1955.

\bibitem{Kohonen1982}
Teuvo Kohonen.
\newblock {Self-Organized Formation of Topologically Correct Feature Maps}.
\newblock {\em Biological Cybernetics}, 43:59--69, 1982.

\bibitem{Kohonen2013}
Teuvo Kohonen.
\newblock Essentials of the self-organizing map.
\newblock {\em Neural Networks}, 37:52--65, 2013.

\bibitem{Strong2011}
Grant Strong and Minglun Gong.
\newblock Data organization and visualization using self-sorting map.
\newblock In Stephen Brooks and Pourang Irani, editors, {\em Graphics Interface}, pages 199--206. Canadian Human-Computer Communications Society, 2011.

\bibitem{Strong2014}
Grant Strong and Minglun Gong.
\newblock Self-sorting map: An efficient algorithm for presenting multimedia data in structured layouts.
\newblock {\em IEEE Trans. Multim.}, 16(4):1045--1058, 2014.

\bibitem{Quadrianto2008}
Novi Quadrianto, Le~Song, and Alexander~J. Smola.
\newblock Kernelized sorting.
\newblock In Daphne Koller, Dale Schuurmans, Yoshua Bengio, and Léon Bottou, editors, {\em NIPS}, pages 1289--1296. Curran Associates, Inc., 2008.

\bibitem{Djuric2012}
Nemanja Djuric, Mihajlo Grbovic, and Slobodan Vucetic.
\newblock Convex kernelized sorting.
\newblock In Jörg Hoffmann and Bart Selman, editors, {\em Proc. AAAI Conference on Artificial Intelligence}, Toronto, Canada, 2012. AAAI Press.

\bibitem{Beckman1957}
M.~Beckman and T.C. Koopmans.
\newblock Assignment problems and the location of economic activities.
\newblock {\em Econometrica}, 25:53--76, 1957.

\bibitem{Fried2015}
Ohad Fried, Stephen DiVerdi, M.~Halber, E.~Sizikova, and Adam Finkelstein.
\newblock Isomatch: Creating informative grid layouts.
\newblock {\em Comput. Graph. Forum}, 34(2):155--166, 2015.

\bibitem{Tenenbaum2000}
Joshua~B. Tenenbaum, Vin de~Silva, and John~C. Langford.
\newblock A global geometric framework for nonlinear dimensionality reduction.
\newblock {\em Science}, 290:2319 -- 2323, 2000.

\bibitem{Kuhn1955}
H.W. Kuhn.
\newblock The hungarian method for the assignment problem.
\newblock {\em Naval research logistics quarterly}, 2(1-2):83--97, 1955.

\bibitem{Hilasaca2021}
Gladys~M. Hilasaca, Wilson~E. Marcílio-Jr, Danilo~M. Eler, Rafael~M. Martins, and Fernando~V. Paulovich.
\newblock Overlap removal of dimensionality reduction scatterplot layouts, 2021.

\bibitem{Jonker1987}
Roy Jonker and A.~Volgenant.
\newblock A shortest augmenting path algorithm for dense and sparse linear assignment problems.
\newblock {\em Computing}, 38(4):325--340, 1987.

\bibitem{46645}
Gonzalo Mena, David Belanger, Scott Linderman, and Jasper~Roland Snoek.
\newblock Learning permutations with gradient descent and the sinkhorn operator.
\newblock In {\em Proc. Int. Conf. on Learning Representations (ICLR)}, Vancouver, Canada, 2018.

\bibitem{Droge2023KissingTF}
Hannah Droge, Zorah Lahner, Yuval Bahat, Onofre Martorell, Felix Heide, and Michael Moller.
\newblock Kissing to find a match: Efficient low-rank permutation representation.
\newblock {\em ArXiv}, abs/2308.13252, 2023.

\bibitem{DBLP:conf/icml/PrilloE20}
Sebastian Prillo and Julian~Martin Eisenschlos.
\newblock Softsort: {A} continuous relaxation for the argsort operator.
\newblock In {\em Proceedings of the 37th International Conference on Machine Learning, {ICML} 2020, 13-18 July 2020, Virtual Event}, volume 119 of {\em Proceedings of Machine Learning Research}, pages 7793--7802. {PMLR}, 2020.

\bibitem{10.1145/3394486.3403103}
Jiancheng Lyu, Shuai Zhang, Yingyong Qi, and Jack Xin.
\newblock Autoshufflenet: Learning permutation matrices via an exact lipschitz continuous penalty in deep convolutional neural networks.
\newblock In {\em Proceedings of the 26th ACM SIGKDD International Conference on Knowledge Discovery \& Data Mining}, KDD '20, page 608–616, New York, NY, USA, 2020. Association for Computing Machinery.

\bibitem{Attention}
Ashish Vaswani, Noam Shazeer, Niki Parmar, Jakob Uszkoreit, Llion Jones, Aidan~N. Gomez, Łukasz Kaiser, and Illia Polosukhin.
\newblock Attention is all you need.
\newblock In {\em Advances in Neural Information Processing Systems 30}. Curran Associates, Inc., 2017.

\bibitem{VIT}
Alexey Dosovitskiy, Lucas Beyer, Alexander Kolesnikov, Dirk Weissenborn, Xiaohua Zhai, Thomas Unterthiner, Mostafa Dehghani, Matthias Minderer, Georg Heigold, Sylvain Gelly, Jakob Uszkoreit, and Neil Houlsby.
\newblock An image is worth 16x16 words: Transformers for image recognition at scale.
\newblock {\em CoRR}, 2020.

\end{thebibliography}

\end{document}